\documentclass[letterpaper]{article} 
\usepackage{aaai24}
\usepackage{times}  
\usepackage{helvet}  
\usepackage{courier}  
\usepackage[hyphens]{url}  
\usepackage{graphicx} 
\usepackage{natbib}  
\usepackage{caption} 
\usepackage{algorithm}
\usepackage{algorithmic}
\usepackage{ascii}
\usepackage{array,multirow}
\usepackage[dvipsnames]{xcolor}

\usepackage{booktabs}
\usepackage{newfloat}
\usepackage{listings}
\DeclareCaptionStyle{ruled}{labelfont=normalfont,labelsep=colon,strut=off} 
\floatstyle{ruled}
\newfloat{listing}{tb}{lst}{}
\floatname{listing}{Listing}
\title{Against All Odds: Overcoming Typology, Script, and Language Confusion in Multilingual Embedding Inversion Attacks}

\author{
    Yiyi Chen,
    Russa Biswas,
    Heather Lent,
    Johannes Bjerva
}
\affiliations{
    Department of Computer Science,
    Aalborg University,
    Copenhagen, Denmark \\
    {yiyic, rubi, hcle, jbjerva}@cs.aau.dk
}

\usepackage{bibentry}

\begin{document}
\nocopyright 

\maketitle

\begin{abstract}
Large Language Models (LLMs) are susceptible to malicious influence by cyber attackers through intrusions such as adversarial, backdoor, and embedding inversion attacks. 
In response, the burgeoning field of LLM Security aims to study and defend against such threats. 
Thus far, the majority of works in this area have focused on monolingual English models, however, emerging research suggests that multilingual LLMs may be more vulnerable to various attacks than their monolingual counterparts. 
While previous work has investigated embedding inversion over a small subset of European languages, it is challenging to extrapolate these findings to languages from different linguistic families and with differing scripts. 
To this end, we explore the security of multilingual LLMs in the context of embedding inversion attacks and investigate cross-lingual and cross-script 
inversion across 20 languages, spanning over 8 language families and 12 scripts.
Our findings indicate that languages written in Arabic script and Cyrillic script are particularly vulnerable to embedding inversion, as are languages within the Indo-Aryan language family. 
We further observe that inversion models tend to suffer from language confusion, sometimes greatly reducing the efficacy of an attack. 
Accordingly, we systematically explore this bottleneck for inversion models, uncovering predictable patterns which could be leveraged by attackers. 
Ultimately, this study aims to further the field's understanding of the outstanding security vulnerabilities facing multilingual LLMs and raise awareness for the languages most at risk of negative impact from these attacks.\footnote{GitHub:~\url{https://github.com/siebeniris/vec2text_exp}, HuggingFace:~\url{https://huggingface.co/yiyic}.
}
\end{abstract}


\section{Introduction}

\begin{figure}[t]
\centering
\includegraphics[width=0.8\columnwidth]{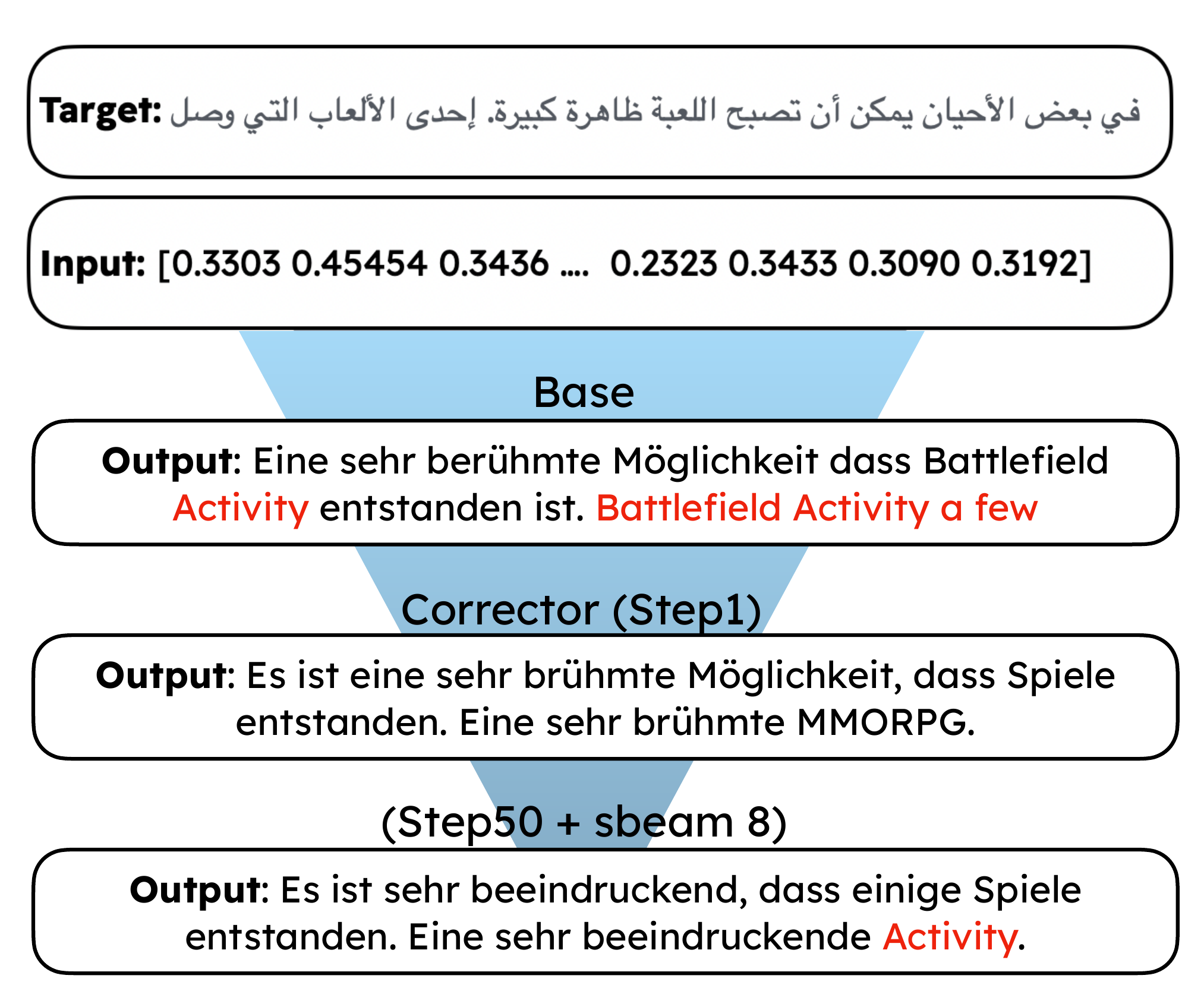}

\caption{Inverting Arabic texts (target) using inversion models trained on multilingual encoder with German (source), showing \textbf{language confusion} across steps (\textcolor{red}{red}).}
\label{fig:top}
\end{figure}

Vector databases store texts encoded into \textit{embeddings}, representing a 1.5 billion USD market globally as of 2023, which is expected to grow to an impressive 4 billion within the next 5 years~\cite{marketresearch23}. 
In more concrete contexts, services like Pinecone and Qdrant,\footnote{\url{https://www.pinecone.io/} and \url{https://qdrant.tech/}} store embeddings for retrieval in applications like retrieval-augmented generation (RAG)~\citep{lewis2020retrieval}. In RAG, user queries are embedded via LLMs and sent to a vector database for semantic search, while large-scale embeddings of text data are also stored for semantic search.
The security of this technology is increasingly critical for providers and customers depending on vector databases for their businesses. 
Presently, vector databases are known to have one outstanding security vulnerability: text embeddings can be reverted into original text, even without prior knowledge of the embedding model, thus constituting a major threat to privacy \cite{10.1145/3372297.3417270, DBLP:journals/corr/abs-2109-10104, hayet-etal-2022-invernet, morris-etal-2023-text, li-etal-2023-sentence, chen2024text}. In other words, text embeddings are no more secure than plain text \cite{morris-etal-2023-text}. 
A malicious actor needs only to intercept the vectors in order to carry out an embedding inversion attack, whereby potentially sensitive information is leaked.

As with most NLP problems, large English datasets have served as the primary focus of research on embedding inversion attacks.
Emerging work, however, highlights how the field's over-reliance on English comes at a grave cost to the general security of Large Language Models (LLMs). 
For example,~\citet{yong2024lowresource} demonstrate how ChatGPT's built-in safety features can be easily by-passed by prompting in lower-resourced languages, while OpenAI's defences function as expected for English and other highly-resourced languages. 
Meanwhile,~\citet{wang2024backdoor} show how low-resource languages can be weaponized to poison models in backdoor attacks against machine translation systems.
In the context of embedding inversion attacks, security may also be compromised by failing to consider a diverse range of languages, especially as vector database providers may leverage multilingual LLMs to serve clients around the globe.
\citet{chen2024text} find that multilingual LLMs are more vulnerable to embedding inversion attacks than their monolingual counterparts. 
However, their study includes only four highly-resourced European languages, restricted to Latin script. 
Conclusions drawn over a small set of relatively homogeneous languages cannot be trivially extended to massively multilingual LLMs, thus stressing the need for
investigations of embedding inversion attacks with languages of diverse linguistic families and varying scripts.  

In this work, we explore the nuances of embedding inversion attacks against multilingual models across 20 languages from 8 language families written in 12 scripts.
We find that the success of inversion attacks is largely influenced by combinations of language family and script, ultimately showing that some languages are more vulnerable to attacks than others. 
Beyond elucidating threats facing multilingual encoders for vector databases, 
we analyze a common hindrance to the performance of embedding inversion:~\textit{language confusion}, 
a term recently coined by~\citet{marchisio2024understanding}, defined as a limitation of 
LLMs where the models are often unable to consistently generate text in the user's desired language, or the appropriate language given the context.
Fig.~\ref{fig:top} illustrates different vulnerability levels in inversion attacks and language confusion. Specifically, Arabic textual embeddings (input) are inverted by a model trained on German embeddings, with the final output producing a mostly coherent German sentence containing an English term.
Language confusion is understood as an ``erroneous behavior'' across the whole of NLP, reported in LLM instruction tuning and is detrimental to generation performance in terms of intelligibility in general~\citep{marchisio2024understanding}.
We show that language confusion can be predicted and controlled for, unfortunately to the benefit of would-be attackers. 


\paragraph{Our contributions:} 

\begin{enumerate}
\item We explore the vulnerability of monolingual versus multilingual LLMs, including a wide variety of languages, and LLMs of differing architectures and training objectives, such that results can be better generalizable for modern massively multilingual LLMs.

\item We investigate embedding inversion attacks across 8 language families, comparing within and across-family and find that the Indo-Aryan family was the most vulnerable. 
Notably, inverting Punjabi texts produces meaningful sentences only in the in-family training setting compared to training it independently.

\item We examine attacks across 12 scripts, finding that scripts differ in vulnerability to inversion attacks. Our findings indicate that Arabic and Cyrillic scripts were the most susceptible to attack. 
In-script training boosts Urdu inversion attack performance, doubling the BLEU score.

\item We identify language confusion as a bottleneck to cross-lingual inversion attacks. Through systematic investigation uncovers that language confusion can be predicted using various language characteristics, such as script and script-directionality. Moreover, the prediction of language confusion can directly aid attack performance.
\end{enumerate}

\section{Related Work}

\paragraph{Textual Embedding Inversion}
Inversion attacks on text embeddings have increased rapidly, with attack accuracy approaching 100\%~\cite{DBLP:journals/corr/abs-2109-10104, hayet-etal-2022-invernet, morris-etal-2023-text, li-etal-2023-sentence} whereas previously only 50--70\% of tokens were achievable~\cite{10.1145/3372297.3417270}.
However, progress has mainly focused on English, leaving the security of multilingual embedding models largely unexplored. The first study to explore embedding inversion attacks in a multilingual context is limited to European languages and Latin script, highlighting a significant gap in LLM security~\cite{chen2024text}. Tokenization methods in multilingual LLMs disproportionately allocate subwords to Latin script~\cite{wu2020all}, suggesting that embedding inversion attack success may differ across languages depending on script and language.
Finally, another notable limitation of the study by \citet{chen2024text} is the reliance on parallel data. As LLMs are known to memorize information \cite{DBLP:journals/corr/ShokriSS16, DBLP:journals/corr/abs-1802-08232, Nasr_2019}, studies on exclusively parallel text preclude the opportunity to study information leakage across languages \cite{ki-etal-2024-mitigating}. As a consequence, the field's threat assessment of multilingual LLM's may not capture the full severity.

\noindent\textbf{Cross-Lingual and Cross-Script Transfer} The majority of the world's languages lack sufficient data for NLP applications \cite{joshi-etal-2020-state, blasi-etal-2022-systematic, ranathunga-de-silva-2022-languages}. 
Cross-lingual transfer offers a solution to this issue, as models trained on resource-rich languages can be applied to low-resource languages \cite{conneau-etal-2020-unsupervised, pmlr-v119-hu20b}. 
While some studies suggest that cross-lingual transfer is more effective within the same language families, other work points to sub-word evenness (SuE) as a more critical factor in successful transfer \cite{pelloni-etal-2022-subword}. 
Furthermore, transferring knowledge to languages with unknown scripts has proven exceedingly difficult \cite{anastasopoulos-neubig-2019-pushing}, prompting exploration of transliteration to mitigate script-related challenges \cite{hermjakob-etal-2018-box, murikinati-etal-2020-transliteration}. 
The implications of cross-lingual and cross-script transfer for LLM security warrant further investigation.

\noindent\textbf{Language Confusion}
Language confusion is a significant limitation of LLMs, where a model cannot consistently generate text in the desired language or appropriate language given the context~\cite{marchisio2024understanding}.
This can result in full-response confusion, line-level confusion, and word-level confusion~\cite{vu2022overcoming,marchisio2024understanding}. In embedding inversion attacks, \citet{chen2024text} attributes language confusion to using a multilingual black-box encoder. 
However, in this work, we find that language confusion can occur when with \textit{both} a multilingual and monolingual black-box encoder.
Language confusion as a vulnerability is also supported by recent work, whereby it has already been proven an effective weapon for jailbreaking state-of-the-art LLMs~\citep{song2024multilingual}.

\section{Methodology}\label{sec:methodology}
In this work, we consider a scenario where a malicious attacker has illegitimate access to stolen embeddings and API access to the black-box encoder. 
To gauge the vulnerability of multilingual models against black-box inversion embedding attacks, we implement inversion attacks on a diverse setting of languages across language families and scripts.
The attack scenario is formally defined as follows: given a text sequence $x$ from a dataset $D$, and a black-box encoder $\phi$, an external inversion model $\psi$ aims to recover the text $x$ from its embeddings $\phi(x)$. 
The objective can be formalized as such that:
\begin{equation}
    \psi(\phi(x)) \approx \phi^{-1}(\phi(x))= x.
\end{equation}
Since sentence embeddings result from pooling token representations, the mapping \( \phi(x) \) is non-injective, making it impossible for the attacker model \( \psi \) to precisely invert \( \phi \)~\citep{li-etal-2023-sentence}.
To learn the approximation of $\psi$, 
we learn a distribution $p(x|e;\theta)$ from a given dataset to invert $e=\phi(x)$ via maximum likelihood~\citep{morris-etal-2023-text}:
\begin{equation}
    \theta = \arg\max_{\hat{\theta}} E_{x \sim D} \left[ p(x \mid \phi(x); \hat{\theta}) \right].
\end{equation}
This forms a text-to-text generation task, resulting in the trained \textbf{base model}. 
However, this approach alone is insufficient.
To improve the inversion performance, a corrector model is implemented, where $\phi$ is queried using the output of the base model, and new \textit{hypothesis embeddings} $\hat{e}^{(t)}=\phi(x^{(t)})$ is computed in iteration $t$ with which a new correction text $x^{(t+1)}$ is generated. 
The corrector aims to find text $\hat{x}$ that maximizes cosine similarity with the target embedding $\phi(x)$ : $\hat{x} = \arg\max_x \cos(\phi(x), e)$. 
At step $(t+1)$, the model concatenates the previous output $\hat{x}^{(t)}$, hypothesis embeddings $\hat{e}^{(t)}$, and target embeddings $\hat{e}^{t}$  as in the equation ~(\ref{eq3}) in the methodology section.
The model is defined by marginalizing over intermediate hypotheses~\citep{morris-etal-2023-text}:

\begin{equation}\label{eq3}
    p(x^{(t+1)}|e) =  \sum_{x^{(t)}} p(x^{(t)}|e)p(x^{(t+1)}|e,x^{(t)},\hat{e}^{(t)})
\end{equation}
We train a \textbf{corrector} model (step1) where $x^{(1)}$ is the correction of $x^{(0)}$, and apply further steps with beam search.
Results are reported for the base model, corrector (step1), and after 50 steps with beam search (width 8).


\paragraph{Language Models} Previous studies have explored language models $\phi$, both
 monolingual such as ~\textsc{gtr}
 and multilingual such as multilingual-e5 (\textsc{me5}) 
but the backbone of $\psi$ is the encoder-decoder T5~\citep{raffel2020exploring}, pre-trained on four Germanic and Romance languages in Latin script, i.e., English, German, Spanish and French.
To explore languages across diverse settings of scripts, we employ multilingual T5 (\textsc{mT5})
as the external encoder-decoder to train the inversion model, pre-trained in 102 languages, including the above languages except Meadow Mari.
\textsc{me5} is also used as our multilingual language black-box encoder, to experiment across settings. The vulnerability between monolingual and multilingual language models is also compared by
adopting \textsc{gtr} as $\phi$ to train the inversion model in German, 
\textsc{alephbert}~\citep{seker-etal-2022-alephbert}
, Hebrew,
and \textsc{text2vec} Chinese. 
This work uses LLMs sourced from Hugging Face (HF).\footnote{HF: sentence-transformers/gtr-t5-base, intfloat/multilingual-e5-base, google/mt5-base, imvladikon/sentence-transformers-alephbert, shibing624/text2vec-base-chinese-paraphrase }.

\paragraph{Sentence Representations}
Prior research shows that BERT’s intermediate layers capture a hierarchy of linguistic information, with surface features in lower layers, syntactic features in the middle, and semantic features in the upper layers ~\citep{jawahar-etal-2019-bert}. 
~\citet{huang-etal-2021-whiteningbert-easy} find that averaging the first and last layers ($s^{[1,12]}=\frac{1}{2}(s^1 + s^{12})$) outperforms the [CLS] token and other layer combinations in downstream tasks like sentence semantics. 
We investigate the existing methods that use last hidden state, mean of all layers, and [CLS] in inversion tasks~\citep{morris-etal-2023-text, chen2024text, zhuang2024understanding} and extend it with $s^{[1,12]}$, revealing that $s^{[1,12]}$ delivers superior attack performance, which is used for model training and evaluation in this paper.



\paragraph{Languages} To explore the scenarios of multi-script and multi-family
inversion attacks, we curate languages written in diverse scripts in several language families. 
Eventually, we simulate the inversion attacks in 20 languages across 8 families and 12 scripts.
We train inversion models and evaluate languages and the combinations of which, 
including 12 languages
\textbf{Arabic} (Semitic; Latn), 
\textbf{Urdu} (Indo-Aryan; Arab), \textbf{Kazakh} (Turkic; Cyrl),
\textbf{Mongolian} (Mongolic, Cyrl), \textbf{Hindi} (Indo-Aryan; Deva), \textbf{Gujarati} (Indo-Aryan; Gujr), \textbf{Punjabi} (Indo-Aryan; Guru), \textbf{Chinese} (Sino-Tibetan; Hani), \textbf{Hebrew} (Semitic; Hebr), \textbf{Japanese} (Japonnic; Jpan), \textbf{German} (Germanic; Latn), and \textbf{Turkish} (Turkic; Latn). 
In addition, we evaluate on another languages \textbf{Amharic} (Semitic; Ethi), \textbf{Sinhala} (Indo-Aryan; Sinh), \textbf{Korean} (Koreanic; Hang), \textbf{Finnish} (Uralic; Latn), \textbf{Hungarian} (Uralic; Latn), \textbf{Yiddish} (Germanic; Hebr), \textbf{Maltese} (Semitic; Latn), and \textbf{Meadow Mari} (Uralic; Cyrl).\footnote{Language name (Language family, ISO 15924 script code). Full descriptions of languages are in SUPPL. doc.}

\paragraph{Dataset} We randomly sample clean and deduplicated data from CulturaX~\citep{nguyen-etal-2024-culturax}, a trillion token dataset in 167 languages. 
Due to data size limitations for certain languages,  such as Indo-Aryan languages, i.e., Hindi, Gujarati, Punjabi and Urdu, we curate 600K samples for each language for training, while for other languages there are 1M samples for each train set, i.e., Arabic, Kazakh, Mongolian, Chinese, Hebrew, Japanese, German and Turkish.
We consider 500 samples for each language in all the above-mentioned 20 languages for evaluation.

\paragraph{Experimental Setup}
Each inversion model consists of a base and a corrector, each trained for 100 epochs with a learning rate of $2e-5$, epsilon of $1e-6$, and 1000 warm-up steps on a constant schedule. 
Models use a batch size of 256 on data with 32 tokens and are trained on 4 AMD MI250 GPUs and 56 CPU cores using distributed training. 
The slowest model takes 4 days to complete in these settings. 

\noindent[\textbf{Baselines}] For each of the above-mentioned 12 languages,  we train and evaluate inversion models using the corresponding data (ref. Table~\ref{tab:baseline-results}).

\noindent[\textbf{In-Script}] To explore inversion attacks within scripts, we train and evaluate on language pairs written in Arabic, Cyrillic, Latin and Hani-Jpan scripts\footnote{Due to the overlapping of Kanji characters between Chinese and Japanese, we experiment on this combination.} (ref. Table~\ref{tab:in-script-results}). 

\noindent[\textbf{In-Family}] Moreover, to better understand the transfer learning within language families in inversion attacks, we train languages in pairs from three language families, i.e., Semitic, Turkic and Indo-Aryan (ref. Table~\ref{tab:family-inversion}). 

\noindent[\textbf{Control}] As a control group, we train the combinations of language pairs from Turkic and Indo-Aryan language families (ref. Table~\ref{tab:random-inversion}). 

\noindent The same data samples used in the baseline models are applied to the in-script and in-family inversion model training. 
For the control group, we matched the number of samples to the smallest dataset.


\paragraph{Word Tokenization and Evaluation}
To measure the matching words between the target and inverted texts, 
we use word tokenizers according to the languages,
Specifically, jieba\footnote{\url{https://github.com/fxsjy/jieba}} (Chinese), 
Hebrew Tokenizer\footnote{\url{https://github.com/YontiLevin/Hebrew-Tokenizer}}  (Hebrew), 
Kiwipiepy~\citep{kiwi-korean} (Korean),
fugashi~\citep{mccann-2020-fugashi} (Japanese), 
and NLTK Word Tokenizer~\citep{bird-loper-2004-nltk} for other languages.
For evaluation, we use word-matching metrics such as BLEU~\citep{post-2018-call}, ROUGE~\citep{lin-2004-rouge}, and Token F1 (TF1), which calculates the multi-class F1 score between predicted and actual tokens. And, cosine similarity (COS) is employed to compare the true and hypothesis embeddings.







\paragraph{Language Confusion}
While previous work in language confusion is directed at prompting in diverse languages using instruct models~\citep{marchisio2024understanding}, our experiment is in the setup, where an inversion model inverts the language embeddings in diverse languages extracted from a black-box encoder. 
Suppose an inversion model is trained on a set of $n\in \mathbf{N}$ languages $L_s=\{l_1, .., l_n\}$, and the text embeddings to be inverted are in a set of $m\in \mathbf{N}$  languages $L_t=\{l_1, .., l_m\}$.
In the context of textual embedding inversion, we investigate language confusion in two settings: \textbf{a)} \textbf{Monolingual generation}, where the inversion model inverts the language embeddings in the languages included in training, i.e., $L_t \subseteq L_s $; \textbf{b)} \textbf{Cross-lingual generation}, where the languages of the target language embeddings are not included in the training data of the inversion model, i.e., $L_t \not\subseteq L_s$ and $L_t \cap L_s = \emptyset $.

To model and predict patterns in language confusion, we leverage the following aspects in the context of the inversion attacks: 
i) Stages of evaluating inversion attacks, i.e., base, corrector (step1) and (Step 50 + sbeam 8);
ii) Language Characteristics, such as \textbf{Langauge Script} (S), \textbf{Language Family} (F), \textbf{Directionality of the Script} (LR)\footnote{Directionality of scripts can include Left-to-Right (LTR), Right-to-Left (RTL), Top-to-Bottom and Bottom-to-Top. In this work, we focus on the scripts written in directions LTR and RTL.}, and \textbf{Word Order of Subject, Object and Verb}(WO)~\citep{wals};
iii) COS;
iv) Comparison of Languages in \textbf{Train Data} (T) and Evaluation Data with (i), (ii) and (iii).

\begin{figure}[ht!]
\centering
\includegraphics[width=0.95\columnwidth]{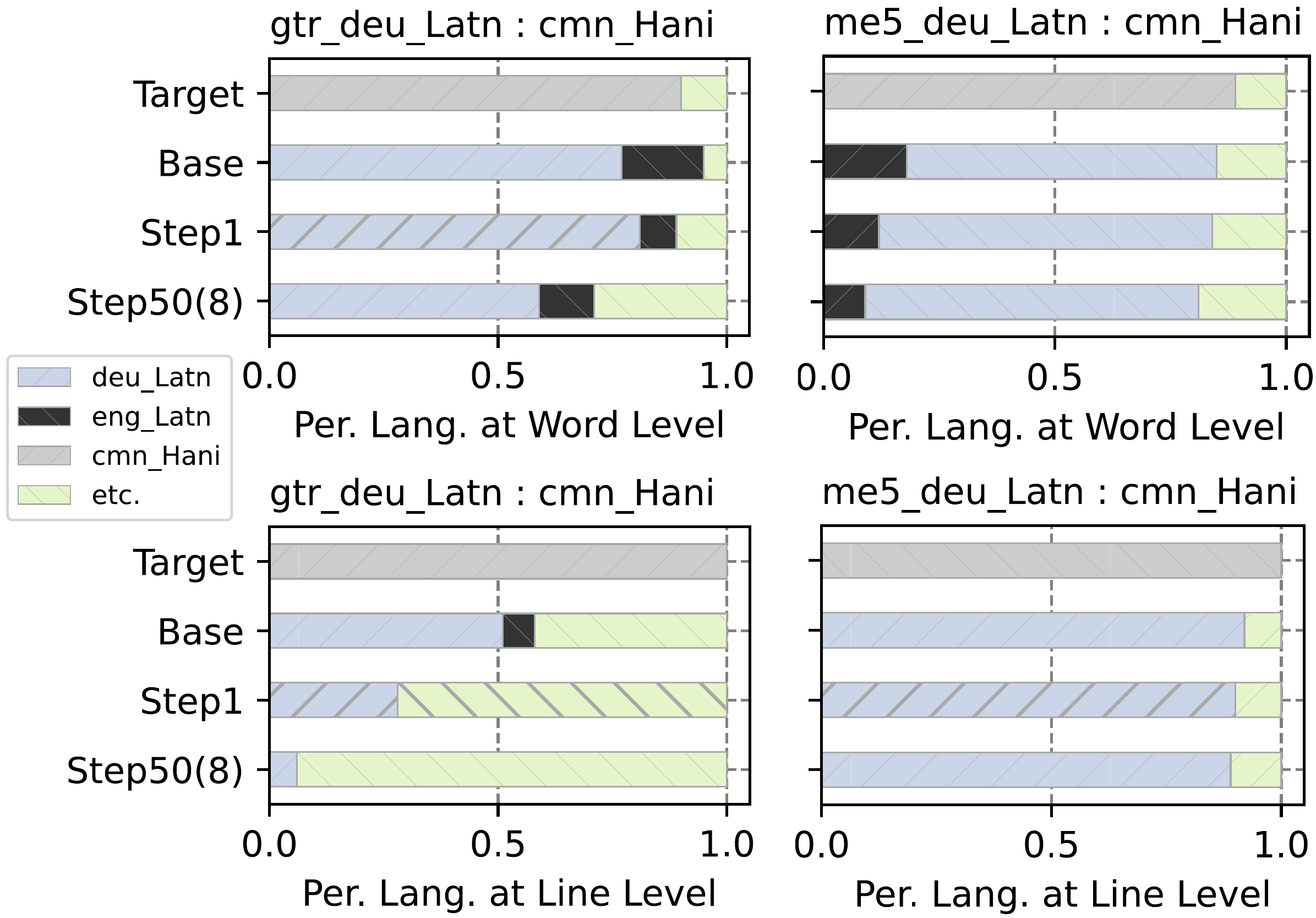}
\caption{Language Confusion of reconstructing Chinese (Hani) texts using inversion models trained on monolingual (left) and multilingual (right) encoders with German (Latn) at Word Level (top) and  Line Level (bottom).}
\label{fig:language_confusion_mono_multi_levels}
\end{figure}

As shown in Fig.~\ref{fig:top}, language confusion can happen at different levels, e.g., word level and line level. Because of loan words and code-switching as such, a mixture of languages can be contained in different levels of texts. To detect different levels of language confusion, we use off-the-shelf language identification (LID) tools.\footnote{\url{https://github.com/pemistahl/lingua-py} and \url{https://github.com/zafercavdar/fasttext-langdetect}}
We define~\textbf{Line-level Language Confusion} under the context of inversion security as the \textit{probabilities} of the detected languages in which the inverted texts are decoded at line level, while~\textbf{Word-level Language Confusion} is defined as such at word level. Using the language detector, we define the language space as the set including all the above-mentioned 20 languages and others (etc.).
As illustrated in Fig.~\ref{fig:language_confusion_mono_multi_levels}, using the inversion model trained on monolingual and multilingual embeddings with German to invert Chinese texts, during different stages, the embeddings in Chinese are inverted into texts in different proportions of English, German and other languages.

After removing missing values, we have 170 samples for monolingual LLMs as $\phi$ and 1789 samples for multilingual LLMs, across monolingual and cross-lingual generation settings. We split the data into 80\% for training and 20\% for testing and use Random Forest\footnote{\url{https://scikit-learn.org/stable/modules/generated/sklearn.ensemble.RandomForestRegressor.html}} to train a regression model. This model predicts the probabilities of the languages in which the inverted embeddings are decoded based on the features described earlier (see details in SUPPL.).
Our goal is to identify which features are more conducive to predicting language identities and to determine if there is a pattern between monolingual and multilingual LLMs.

\section{Results on Inversion Attacks}

\begin{table}[!ht]
   \centering
  \resizebox{0.4\textwidth}{!}{
  \begin{tabular}{l|c|c|c|c|c|c}
     \toprule
  \textbf{Language (Script)}   & \textbf{\#Tok.} & \textbf{\#Pred. Tok.} & \textbf{Token F1} & \textbf{BLEU} & \textbf{ROUGE} & \textbf{COS}  \\ 
    \midrule
   
     \textbf{Arabic (Arab)} & ~ & ~ & ~ & ~ & ~ & ~ \\ 
       Base & 13.48 & 13.27 & 40.06 & 17.31 & 92.93 & \underline{0.9310} \\ 
        Corrector (Step1) & 13.48 & 13.02 & 43.66 & 19.12 & 92.96 & 0.9292 \\ 
        (Step 50+sbeam8) & 13.48 & 13.17 & \underline{48.18} & \textbf{22.93} & 94.15 & 0.9232 \\ 
\midrule

        \textbf{Urdu (Arab)} & ~ & ~ & ~ & ~ & ~ & ~ \\ 
        Base & 15.03 & 14.77 & 49.71 & 20.86 & 78.57 & 0.9612 \\
        Corrector (Step1) & 15.03 & 14.72 & 51.64 & 22.76 & 79.27 & \underline{0.9736} \\
        (Step 50+sbeam8) & 15.03 & 14.85 & \underline{54.66} & \textbf{25.04} & 80.19 & 0.9712 \\          
    \midrule
    
    \textbf{Kazakh (Cyrl)} & ~ & ~ & ~ & ~ & ~ & ~ \\ 
        Base & 13.35 & 13.16 & 53.12 & 30.69 & 86.39 & 0.9712 \\ 
         Corrector (Step1)  & 13.35 & 13.12 &  57.46 &  34.62 & 87.29 & 0.9794 \\
       (Step 50+sbeam8) & 	13.35 &	13.21	& \underline{65.74} & 	\textbf{44.50}	& 89.44 &	\underline{0.9860}	\\												
        \midrule

          \textbf{Mongolian (Cyrl)} & ~ & ~ & ~ & ~ & ~ & ~ \\ 
        Base & 12.00 & 11.74 & 50.01 & 28.21 & 82.36 & \underline{0.9818} \\ 
         Corrector (Step1)  & 12.00 & 11.70 & 55.36 &  32.96 & 85.00 & 0.9523 \\ 
         (Step 50+sbeam8) &  12.00 & 11.85	& \underline{64.67}	& \textbf{43.63} & 88.58 & 0.9742\\														
        \midrule
        
        \textbf{Hindi (Deva)} & ~ & ~ & ~ & ~ & ~ & ~ \\ 
        Base & 15.21 & 14.98 & 50.98 & 19.10 & 73.40 & \underline{0.9576} \\ 
        Corrector (Step1) & 15.21 & 14.77 & 52.78 & 20.06 & 74.25 & 0.9443 \\ 
        (Step 50+sbeam8)  & 15.21 & 14.77 & \underline{55.79} & \textbf{22.29} & 76.00 & 0.9197 \\ 
        
        \midrule

         \textbf{Gujarati (Gujr)} & ~ & ~ & ~ & ~ & ~ & ~ \\ 
        Base & 11.07 & 10.85 & 43.19 & 18.74 & 87.57 & 0.9787 \\ 
        Corrector (Step1) & 11.07 & 10.71 & 45.51 & 20.15 & 87.65 & 0.9856 \\ 
        (Step 50+sbeam8) & 11.07 & 10.81 &\underline{48.17} & \textbf{22.35} & 88.43 & \underline{0.9893} \\ 
        \midrule

     \textbf{Punjabi (Guru)} & ~ & ~ & ~ & ~ & ~ & ~ \\ 
        Base & 11.07 & 10.97 & \underline{59.65} & \textbf{31.10} & 91.39 & \underline{0.9882} \\ 
        Corrector (Step1) & 11.07 & 21 & 0.03 & 0.43 & 0 & 0.7355 \\ 
         (Step 50+sbeam8) & 	11.07	& 21.00	&0.03	&0 &	0 &	0.7190	 \\
        \midrule

        \textbf{Chinese (Hani)} & ~ & ~ & ~ & ~ & ~ & ~ \\ 
        Base & 5.81 & 6.06 & 27.84 & 16.73 & 81.82 & 0.9483 \\ 
        Corrector (Step1) & 5.81 & 5.88 & 28.33 & 17.12 & 82.29 & \underline{0.9701} \\ 
        (Step 50+sbeam8) & 5.81 & 5.74 & \underline{29.2} & \textbf{17.50} & 83.35 & 0.9543 \\ 
    \midrule
    
     \textbf{Hebrew (Hebr)} & ~ & ~ & ~ & ~ & ~ & ~ \\ 
       Base & 13.94 & 13.67 & 38.89 & 14.84 & 91.63 & \underline{0.9646} \\ 
       Corrector (Step1) & 13.94 & 13.50 & 41.81 & 16.32 & 92.30 & 0.9588 \\ 
       (Step 50+sbeam8)  & 13.94 & 13.68 & \underline{45.1} & 
       \textbf{18.77} & 93.04 & 0.9477 \\ 
       \midrule
       
  \textbf{Japanese (Jpan)} & ~ & ~ & ~ & ~ & ~ & ~ \\ 
        Base & 2.93 & 3.03 & 9.47 & 9.31 & 80.08 & 0.9496 \\ 
         Corrector (Step1) & 2.93 & 2.94 & 9.46 & 8.93 & 80.41 & 0.9576 \\ 
        (Step 50+sbeam8)  &  2.93 & 2.87 & \underline{9.61}	 & \textbf{8.60} & 81.66 & \underline{0.9630} \\
    
        \midrule
        \textbf{German (Latn)} & ~ & ~ & ~ & ~ & ~ & ~ \\ 
        Base & 17.26 & 16.76 & 55.46 & 25.88 & 57.22 & 0.9460 \\ 
        Corrector (Step1) & 17.26 & 16.85 & 56.69 & 27.35 & 58.78 & 0.9563 \\ 
        (Step 50+sbeam8) & 17.26 & 16.80 & \underline{61.85} & \textbf{31.35} & 64.62 & \underline{0.9982} \\ 

        \midrule

         \textbf{Turkish (Latn)} & ~ & ~ & ~ & ~ & ~ & ~ \\ 
        Base & 14.66 & 14.36 & 44.05 & 18.15 & 49.40 & 0.9498 \\ 
        Corrector (Step1) & 14.66 & 14.26 & 46.33 & 20.62 & 52.51 & \underline{0.9650} \\ 
        (Step 50+sbeam8) & 14.66 & 14.30 & \underline{50.38} & \textbf{24.51} & 57.51 & 0.9309 \\ 

        \bottomrule
    \end{tabular}}
    \caption{Text Reconstruction in Multiple Languages, sorted by the alphabetic order of language scripts. For each language, the best performing TF1 and COS are \underline{underlined}, and the best BLEU is \textbf{bolded}. }
    \label{tab:baseline-results}
\end{table}

\paragraph{Baselines}
We train and evaluate inversion models on \textsc{me5} embeddings across 12 languages. 
As detailed in Table~\ref{tab:baseline-results}, the performance of attacking textual embeddings improves progressively from the base model to the corrector model (Step 1), and further with 50 steps and a beam search width of 8. 
An exception is observed with inverting Punjabi text embeddings, which achieve the highest performance in BLEU and TF1 at the base model but deteriorate significantly when using the corrector.
Overall, multilingual embeddings are more susceptible to attacks in languages written in Cyrillic compared to those in Arabic, Latin, or Hebrew scripts, particularly in Haqniqdoq and Japanese scripts. 
Despite being trained on a smaller dataset, Indo-Aryan languages are more vulnerable than Chinese, Japanese, and Hebrew. 
This finding aligns with previous work by \citet{wang-etal-2024-languages}, which found that Hindi and Bengali suffered from more safety issues than other languages, in the context of adversarial attacks.

\begin{table}[ht!]
\centering
  \resizebox{\columnwidth}{!}{
\begin{tabular}{l|rr|rr|rr|rr}
    
     \toprule
  
\textbf{Language (Script)}  &\multicolumn{2}{c}{\textbf{Token F1}}  & \multicolumn{2}{c}{\textbf{BLEU}} &  \multicolumn{2}{c}{\textbf{ROUGE}}  &  \multicolumn{2}{c}{\textbf{COS}}  \\ 
    & \textsc{mono} & \textsc{multi}  & \textsc{mono} & \textsc{multi}  &  \textsc{mono} & \textsc{multi} & \textsc{mono} & \textsc{multi} \\
\midrule

    \textbf{Chinese (Hani)} & ~ & ~ & ~ & ~ & ~ & ~ & ~ & ~ \\ 
        Base & 21.87 & 27.84 & 6.53 & 16.73 & 6.82 & 81.82 & 0.8723 & 0.9484 \\ 
        Corrector (Step 1) & 22.05 & 28.33 & 6.19 & 17.12 & 7.04 & 82.29 & 0.8548 & \textbf{0.9701} \\ 
        (Step 50+sbeam 8) & 21.75 & \textbf{29.2} & 6.01 & \textbf{17.50} & 7.51 & \textbf{83.35} & 0.8576 & 0.9543 \\
        
        \midrule

    \textbf{Hebrew (Hebr)} & ~ & ~ & ~ & ~ & ~ & ~ & ~ & ~ \\ 
        Base & 34.17 & 38.89 & 8.99 & 14.84 & 2.79 & 91.63 & 0.8601 & \textbf{0.9646} \\ 
        Corrector (Step 1) & 36.27 & 41.81 & 9.95 & 16.32 & 2.46 & 92.30 & 0.8773 & 0.9588 \\ 
        (Step 50+sbeam 8) & 37.98 & \textbf{45.1} & 10.56 & \textbf{18.77} & 2.98 & \textbf{93.04} & 0.9018 & 0.9477 \\ 
        \midrule
        
       \textbf{German (Latn)} & ~ & ~ & ~ & ~ & ~ & ~ & ~ & ~ \\ 
        Base & 39.86 & 55.46 & 15.07 & 25.88 & 38.19 & 57.22 & 0.7985 & 0.9460 \\
        Corrector (Step 1) & 39.83 & 56.69 & 14.52 & 27.35 & 38.55 & 58.78 & 0.9083 & 0.9563 \\ 
        (Step 50+sbeam 8) & 42.17 & \textbf{61.85} & 15.54 & \textbf{31.35} & 41.15 & \textbf{64.62} & 0.9718 & \textbf{0.9982} \\

\bottomrule
    \end{tabular}}
    \caption{\textsc{mono} and \textsc{multi} evaluate Text Reconstruction with monolingual embeddings and multilingual embeddings respectively. The best results for metrics are in \textbf{bold}. }
    \label{tab:mono-multi-results}
    
\end{table}

\paragraph{Monolingual vs. Multilingual LLMs}
To compare the vulnerability of monolingual embeddings and multilingual embeddings, we train and evaluate German, Hebrew, and Chinese on both monolingual (\textsc{mono}) and multilingual (\textsc{multi}) LLMs.
While in~\citet{chen2024text}, with the same settings of \textsc{mono} and \textsc{multi}, trained on English samples, the best reconstruction performance on \textsc{mono} excels \textsc{multi} by $14.33\%$ in BLEU and by $5.81\%$ in TF1.
While trained on German samples, inverting \textsc{mono} embeddings under-performs across the metrics compared to \textsc{multi}, as shown in Table~\ref{tab:mono-multi-results}. 
Overall, while the performance improve steadily across languages across the phases of evaluations in both word matching metrics and cosine similarities, \textsc{me5} is more vulnerable compared to \textsc{mono} LLMs with different architectures (see details of the LLMs in SUPPL.). 
Also consistent with the findings from~\citet{chen2024text}, the cosine similarities are uniformly higher for \textsc{multi} than \textsc{mono}. 
Overall, German is more vulnerable in both monolingual and multilingual settings, compared to Chinese and Hebrew.

\begin{table}[t]
   \centering
  \resizebox{\columnwidth}{!}{
  \begin{tabular}{c|l|c|c|c|c|c|c}
    \toprule
      &\textbf{Language (Script)} & \textbf{\# Tok.} & \textbf{\# Pred. Tok.} & \textbf{Token F1} & \textbf{BLEU} & \textbf{ROUGE} & \textbf{COS}  \\ 

      \midrule

    \multirow{8}{*}{\rotatebox[origin=c]{90}{\textbf{Arabic Script}}} 
& \textbf{Arabic} & ~ & ~ & ~ & ~ & ~ & ~ \\ 
       &  Base & 13.48 & 13.37 & 41.91 ($\uparrow$4.62\%) & 17.97 ($\uparrow$3.81\%) & 93.69 & 0.9416 \\ 
      &   Corrector (Step1) & 13.48 & 13.24 & 47.13 ($\uparrow$7.95\%) & 21.48 ($\uparrow$12.34\%) & 94.06 & 0.9332 \\
         &  (Step 50+sbeam8) & 13.48 & 13.30 & \underline{55.58} ($\uparrow$15.36\%) & \textbf{28.73} (\colorbox{Dandelion}{$\uparrow$25.29\%}) & 94.68 & 0.9243 \\

          &  \textbf{Urdu} & ~ & ~ & ~ & ~ & ~ & ~ \\ 
     & Base & 15.03 & 14.93 & 61.39 ($\uparrow$23.50\%) & 31.05 ($\uparrow$48.85\%) & 88.61 & 0.9735 \\ 

      & Corrector (Step1) & 15.03 & 14.83 & 66.7 ($\uparrow$29.16\%)& 36.90 ($\uparrow$62.13\%) & 90.41 & 0.9618 \\ 
           &  (Step 50+sbeam8) & 15.03 & 14.90 & \underline{76.18} ($\uparrow$39.37\%) & \textbf{50.13} (\colorbox{Dandelion}{$\uparrow$100.20\%}) & 92.80 & 0.9832 \\ 

           \midrule

    \multirow{8}{*}{\rotatebox[origin=c]{90}{\textbf{Hani-Jpan Script}}}
& \textbf{Chinese} & ~ & ~ & ~ & ~ & ~ & ~ \\ 
        & Base & 5.81 & 6.08 & 28.79 ($\uparrow$3.41\%) & 17.10 ($\uparrow$2.21\%)& 82.52 & 0.9475 \\ 
         &   Corrector (Step1) & 5.81 &	5.90 & 30 ($\uparrow$5.89\%) &	18.01 ($\uparrow$5.20\%) &	84.82 &	0.9549 \\ 
            &  (Step 50+sbeam8)& 5.81 &	5.71 &	\underline{31.66} ($\uparrow$8.42\%) &	\textbf{19.86} ($\uparrow$13.49\%) &	87.80 &	0.9476 \\ 
            
            & \textbf{Japanese} & ~ & ~ & ~ & ~ & ~ & ~ \\ 
        & Base  & 2.93 & 2.94 & 9.85 ($\uparrow$4.01\%) & 9.41 ($\uparrow$1.07\%) & 80.64 & 0.9535 \\
       &   Corrector (Step1) & 2.93 &	2.84 &	10.63 ($\uparrow$12.37\%)	 &8.99 ($\uparrow$0.67\%) &	83.23 &	0.9698 \\ 
          &  (Step 50+sbeam8) & 2.93 &	2.94 &	\underline{11.58} ($\uparrow$20.50\%) &	\textbf{10.25} ($\uparrow$19.17\%) &	85.61	 &0.9666 \\ 

        \midrule

\multirow{8}{*}{\rotatebox[origin=c]{90}{\textbf{Cyrillic Script}}}
 & \textbf{Kazakh} & ~ & ~ & ~ & ~ & ~ & ~ \\ 

       &   Base & 13.35 & 13.17 & 54.29 ($\uparrow$2.20\%)& 31.39 ($\uparrow$2.28\%) & 87.08 & 0.9682 \\ 
         &   Corrector (Step1) & 13.35 & 13.08 & 61 ($\uparrow$6.16\%)& 37.62 ($\uparrow$8.67\%) & 89.83 & 0.9836 \\ 
       &  (Step 50+sbeam8) & 13.35 & 13.21 & \underline{72.37} ($\uparrow$10.09\%) & \textbf{52.44} (\colorbox{Dandelion}{$\uparrow$17.85\%}) & 92.58 & 0.9863 \\ 
 & \textbf{Mongolian} & ~ & ~ & ~ & ~ & ~ & ~ \\ 
        &   Base & 12.00 & 11.75 & 53.56 ($\uparrow$7.10\%) & 31.63 ($\uparrow$12.12\%) & 85.14 & 0.9716 \\ 
         &   Corrector (Step1) & 12.00 & 11.67 & 59.74 ($\uparrow$7.91\%) & 37.47 ($\uparrow$13.68\%) & 87.15 & 0.9777 \\ 
         &  (Step 50+sbeam8) & 12.00 & 11.82 & \underline{71.96} ($\uparrow$11.27\%) & \textbf{51.62} ($\uparrow$18.31\%) & 91.22 & 0.9985 \\ 

          \midrule

\multirow{8}{*}{\rotatebox[origin=c]{90}{\textbf{Latin Script}}} &  
       \textbf{German} & ~ & ~ & ~ & ~ & ~ & ~ \\ 
       & Base  & 17.26 & 16.90 & 53.56 ($\downarrow$3.43\%) & 24.20 ($\downarrow$6.49\%) & 55.35 & 0.9545 \\ 
          &  Corrector (Step1) & 17.26 & 17.00 & 58.2 ($\uparrow$2.66\%)& 27.44 ($\uparrow$0.33\%) & 60.23 & 0.9594 \\ 
        &  (Step 50+sbeam8) & 17.26 & 16.96 & \underline{68.42} ($\uparrow$10.62\%) & \textbf{37.66} ($\uparrow$20.13\%) & 71.11 & 0.9958 \\ 

      &  \textbf{Turkish} & ~ & ~ & ~ & ~ & ~ & ~ \\ 
        
          &   Base  & 14.66 & 14.47 & 42.96 ($\downarrow$2.47\%) & 17.29 ($\downarrow$4.74\%) & 48.97 & 0.9595 \\ 
        &   Corrector (Step1)  & 14.66 & 14.30 & 50.08 ($\uparrow$8.09\%) & 23.36 ($\uparrow$13.29\%) & 56.26 & 0.9637 \\
         &   (Step 50+sbeam8) & 14.66 & 14.39 & \underline{59.68} ($\uparrow$18.46\%) & \textbf{33.33} ($\uparrow$35.99\%) & 66.35 & 0.9680 \\ 
        
        \bottomrule
        
    \end{tabular}}
    \caption{Multilingual Text Reconstruction within Script. For each language, the best TF1 score is \underline{underlined}, and the best BLEU score is \textbf{bolded}. The \colorbox{Dandelion}{color box} indicates the highest inversion boost compared to the baseline. }
    \label{tab:in-script-results}
\end{table}

\paragraph{Attacking In-script Textual Embeddings}
Since cross-lingual transfer learning is pervasive in multilingual NLP, and previous work has only touched upon Latin script in text embedding inversion, we investigate whether incorporating multiple languages in the same script would boost the inversion attack performance. 
Although the Japanese writing script has a more diverse set of forms (i.e., kanji and kana) compared to Chinese, they share a fair amount of vocabulary in the same script, in kanji and hanzi. 
To minimize the influence of transfer learning due to shared language families, the languages in each script are selected from different language families.
As shown in Table~\ref{tab:in-script-results}, we train the inversion model in Arabic script with Arabic and Urdu, Cyrillic with Kazakh and Mongolian and Latin with German and Turkic, in addition to the Hani-Jpan combination with Chinese and Japanese.
Overall, the inversion attack performances improve across languages when they are trained together with the same-script languages across stages, except in rare cases such as inverting German and Turkish embeddings with the base model. 
Language embeddings in Cyrillic are the most vulnerable compared to other scripts while Japanese and Chinese appear to be less vulnerable than other evaluated languages. Notably, the performance of attacking Urdu text embeddings is boosted notably by $100\%$ in BLEU score and $39.37$\% in TF1 compared to the baseline. 
Arabic, Urdu, Kazakh and Turkish are further investigated in the scenarios of in-family inversion attacks.
Overall, the in-script training improves performance more than in-family training, except for Turkish (ref. Table~\ref{tab:in-script-results} and \ref{tab:family-inversion}).

\begin{table}[t!]
    \centering
    \resizebox{\columnwidth}{!}{
  \begin{tabular}{c|l|c|c|c|c|c|c}
    \toprule
     & \textbf{Language (Script)}& \textbf{\#Tok.} & \textbf{\#Pred. Tok.} & \textbf{Token F1} & \textbf{BLEU} & \textbf{ROUGE} & \textbf{COS} \\ 
     \midrule
      \multirow{8}{*}{\rotatebox[origin=c]{90}{\textbf{Semitic}}} &
        \textbf{Arabic (Arab)} & ~ & ~ & ~ & ~ & ~ & ~ \\ 
       & Base & 13.48 & 13.34 & 42.75 ($\uparrow$6.71\%) & 18.56 ($\uparrow$7.22\%) & 93.64 & 0.9434 \\
        &Corrector (Step1) & 13.48 & 13.14 & 46.46 ($\uparrow$6.41\%) & 20.97 ($\uparrow$9.65\%) & 93.97 & 0.9572 \\ 
       & (Step 50+sbeam8) & 13.48 & 13.25 & \underline{52.27}($\uparrow$8,49\%) & \textbf{25.61}($\uparrow$11.68\%) & 94.35 & 0.9140 \\ 
       
      &  \textbf{Hebrew (Hebr)} & ~ & ~ & ~ & ~ & ~ & ~ \\ 
        & Base & 13.94 & 13.65 & 39.88 ($\uparrow$2.55\%)& 15.02 ($\uparrow$1.18\%)& 93.36 & 0.9583 \\ 
           &Corrector (Step1)  & 13.94 & 13.49 & 43.6 ($\uparrow$4.28\%) & 17.26 ($\uparrow$5.78\%) & 93.87 & 0.9707 \\ 
        & (Step 50+sbeam8)  & 13.94 & 13.57 & \underline{48.4} ($\uparrow$7.32\%) & \textbf{20.30}($\uparrow$ 8.15\%) & 94.06 & 0.9730 \\ 
        \midrule

     \multirow{8}{*}{\rotatebox[origin=c]{90}{\textbf{Turkic}}}
           & \textbf{Kazakh (Cyrl)} & ~ & ~ & ~ & ~ & ~ & ~ \\ 
        & Base & 13.35 & 13.14 & 55.07 ($\uparrow$3.67\%) & 31.62 ($\uparrow$3.02\%) & 87.83 & 0.9755 \\ 
         &Corrector (Step1)  & 13.35 & 13.20 & 61.07 ($\uparrow$6.28\%)& 37.85 ($\uparrow$9.34\%) & 89.82 & 0.9685 \\
          & (Step 50+sbeam8)  & 13.35 & 13.24 & 
          \underline{69.71} ($\uparrow$6.04\%)& \textbf{49.94} ($\uparrow$12.24\%) & 92.56 & 0.9815 \\
     &   \textbf{Turkish (Latn)} & ~ & ~ & ~ & ~ & ~ & ~ \\ 
        & Base & 14.66 & 14.41 & 45.32 ($\uparrow$2.88\%) & 19.31 ($\uparrow$6.40\%) & 51.52 & 0.9399 \\ 
       &Corrector (Step1)  & 14.66 & 14.34 & 51.07 ($\uparrow$10.23\%) & 23.96 ($\uparrow$16.19\%) & 57.22 & 0.9597 \\
        & (Step 50+sbeam8)   & 14.66 & 14.41 & \underline{59.58} ($\uparrow$18.26\%) & \textbf{33.62} (\colorbox{Dandelion}{$\uparrow$37.16\%}) & 66.10 & 0.9683 \\ 
          \midrule

    \multirow{8}{*}{\rotatebox[origin=c]{90}{\textbf{Indo-Aryan}}} &  
        \textbf{Hindi (Deva)} & ~ & ~ & ~ & ~ & ~ & ~ \\ 
        & Base  & 15.21 & 15.02 & 51.07 ($\uparrow$0.18\%) & 19.47 ($\uparrow$1.94\%) & 73.95 & 0.9596 \\
         &Corrector (Step1) & 15.21 & 14.88 & 54.97 ($\uparrow$4.15\%) & 22.72 ($\uparrow$13.24\%) & 75.65 & 0.9733 \\ 
         & (Step 50+sbeam8)  & 15.21 & 15.04 &\underline{62.12} ($\uparrow$11.35\%) & \textbf{28.72} ($\uparrow$28.86\%)& 79.20 & 0.9728 \\ 
         
        & \textbf{Gujarati (Gujr)} & ~ & ~ & ~ & ~ & ~ & ~ \\ 
       & Base & 11.07 & 10.77 & 43.91 ($\uparrow$1.67\%) & 18.99 ($\uparrow$1.32\%)& 88.82 & 0.9824 \\ 
          &Corrector (Step1)  & 11.07 & 10.74 & 48.97 ($\uparrow$7.60\%) & 22.12 ($\uparrow$9.78\%) & 89.52 & 0.9870 \\ 
        & (Step 50+sbeam8)  & 11.07 & 10.89 & \underline{56.08} ($\uparrow$16.42\%) & \textbf{29.54} ($\uparrow$32.17\%) & 91.42 & 0.9925 \\ 
        \midrule 

        \multirow{8}{*}{\rotatebox[origin=c]{90}{\textbf{Indo-Aryan}}}
       & \textbf{Hindi} (Deva) & ~ & ~ & ~ & ~ & ~ & ~ \\
         & Base & 15.21 & 15.03 & 53.21 ($\uparrow$4.37\%) & 20.62 ($\uparrow$7.98\%)& 74.83 & 0.9476 \\ 
         &Corrector (Step1)  & 15.21 & 14.91 & 57.07 ($\uparrow$8.13\%)& 23.73 ($\uparrow$18.30\%) & 77.00 & 0.9389 \\ 
          & (Step 50+sbeam8)  & 15.21 & 14.98 & \underline{62.31} ($\uparrow$11.69\%)& \textbf{28.89} ($\uparrow$29.61\%) & 79.03 & 0.9115 \\ 
          
      & \textbf{Punjabi} (Guru) & ~ & ~ & ~ & ~ & ~ & ~ \\ 
       & Base & 11.07 & 10.90 & 63.6 ($\uparrow$6.62\%)& 35.78 ($\uparrow$15.04\%) & 93.05 & 0.9872 \\ 
       &Corrector (Step1) & 11.07 & 10.87 & 68.1 ($\uparrow$$\uparrow$) & 40.61 ($\uparrow$$\uparrow$) & 93.80 & 0.9903 \\ 
       & (Step 50+sbeam8)  & 11.07 & 10.97 & \underline{75.37} ($\uparrow$$\uparrow$) & \textbf{50.80} ($\uparrow$$\uparrow$) & 95.08 & 0.9967 \\ 
       \midrule

        \multirow{8}{*}{\rotatebox[origin=c]{90}{\textbf{Indo-Aryan}}}
        & \textbf{Gujarati} (Gujr) & ~ & ~ & ~ & ~ & ~ & ~ \\ 
         & Base & 11.07 & 10.81 & 44.1 ($\uparrow$2.11\%)& 19.47 ($\uparrow$3.88\%) & 87.90 & 0.9824 \\ 
         &Corrector (Step1)  & 11.07 & 10.67 & 36.63 ($\downarrow$19.51\%) & 13.49 ($\downarrow$33.06\%) & 85.59 & 0.9838 \\ 
          & (Step 50+sbeam8)  & 11.07 & 10.84 & \underline{50.66} ($\uparrow$5.17\%) & \textbf{24.22} ($\uparrow$8.37\%) & 89.74 & 0.9880 \\ 
          
        & \textbf{Punjabi} (Guru) & ~ & ~ & ~ & ~ & ~ & ~ \\ 
       & Base  & 11.07 & 11.00 & 63.43 ($\uparrow$6.34\%) & 34.64 ($\uparrow$11.39\%) & 93.21 & 0.9883 \\ 
        &Corrector (Step1) & 11.07 & 10.90 & 68.56 ($\uparrow$$\uparrow$) & 41.07 ($\uparrow$$\uparrow$) & 94.39 & 0.9887 \\ 
      & (Step 50+sbeam8) & 11.07 & 10.99 & \underline{76.51} ($\uparrow$$\uparrow$) & \textbf{52.89} (\colorbox{Dandelion}{$\uparrow$$\uparrow$}) & 95.43 & 0.9957 \\ 
      
        \midrule

        \multirow{8}{*}{\rotatebox[origin=c]{90}{\textbf{Indo-Aryan}}}
       & \textbf{Gujarati} (Gujr) & ~ & ~ & ~ & ~ & ~ & ~ \\ 
       & Base & 11.07 & 10.79 & 45.42($\uparrow$5.16\%) & 19.88 ($\uparrow$6.07\%)& 87.62 & 0.9763 \\ 
        &Corrector (Step1) & 11.07 & 10.72 & 49.72 ($\uparrow$9.25\%) & 23.27 ($\uparrow$15.49\%) & 88.52 & 0.9850 \\ 
         & (Step 50+sbeam8) & 11.07 & 10.83 & \underline{56.89} ($\uparrow$18.10\%) & \textbf{30.34} ($\uparrow$35.74\%) & 90.55 & 0.9949 \\ 
       & \textbf{Urdu} (Arab) & ~ & ~ & ~ & ~ & ~ & ~ \\ 
       & Base & 15.03 & 14.84 & 54.63 ($\uparrow$9.90\%) & 24.23 ($\uparrow$16.17\%)& 83.28 & 0.9621 \\ 
         &Corrector (Step1)  & 15.03 & 14.78 & 57.69 ($\uparrow$11.72\%) & 27.35 ($\uparrow$20.16\%) & 85.43 & 0.9709 \\ 
          & (Step 50+sbeam8) & 15.03 & 14.84 & \underline{62.8} ($\uparrow$14.89\%) & \textbf{32.99} ($\uparrow$31.75\%) & 87.40 & 0.9824 \\ 
          
        \midrule

        \multirow{8}{*}{\rotatebox[origin=c]{90}{\textbf{Indo-Aryan}}}
        & \textbf{Hindi} (Deva) & ~ & ~ & ~ & ~ & ~ & ~ \\ 
        & Base  & 15.21 & 15.01 & 51.43 ($\uparrow$0.88\%) & 18.84 ($\downarrow$1.35\%)& 73.20 & 0.9427 \\ 
         &Corrector (Step1) & 15.21 & 14.88 & 55.22 ($\uparrow$4.62\%) & 21.57 ($\uparrow$7.53\%) & 75.84 & 0.9772 \\ 
        & (Step 50+sbeam8) & 15.21 & 14.98 & \underline{61.21} ($\uparrow$9.72\%) & \textbf{27.93} ($\uparrow$25.31\%) & 78.82 & 0.9771 \\ 
        
         & \textbf{Urdu} (Arab) & ~ & ~ & ~ & ~ & ~ & ~ \\ 
        & Base  & 15.03 & 14.75 & 53.32 ($\uparrow$7.26\%) & 23.87 ($\uparrow$14.42\%) & 83.86 & 0.9728 \\ 
         &Corrector (Step1) & 15.03 & 14.87 & 57.19 ($\uparrow$10.75\%) & 26.84 ($\uparrow$17.94\%) & 85.60 & 0.9683 \\ 
         & (Step 50+sbeam8)& 15.03 & 15.01 & \underline{63.15} ($\uparrow$15.53\%) & \textbf{32.81} ($\uparrow$31.03\%) & 87.79 & 0.9831 \\ 
         
         \midrule

    \multirow{8}{*}{\rotatebox[origin=c]{90}{\textbf{Indo-Aryan}}}
        & \textbf{Punjabi} (Guru) & ~ & ~ & ~ & ~ & ~ & ~ \\ 
        & Base  & 11.07 & 10.89 & 60.99 ($\uparrow$2.25\%) & 33.19 ($\uparrow$6.72\%)& 91.76 & 0.9878 \\ 
        &Corrector (Step1) & 11.07 & 10.62 & 38.94 ($\uparrow$$\uparrow$) & 13.98 ($\uparrow$$\uparrow$) & 88.27 & 0.9739 \\ 
       & (Step 50+sbeam8)&	11.07 &	10.89&	\underline{59.99} ($\uparrow$$\uparrow$) &	\textbf{30.86} ($\uparrow$$\uparrow$) & 0.9218 &	0.9878	\\													
        
        & \textbf{Urdu} (Arab) & ~ & ~ & ~ & ~ & ~ & ~ \\ 
         & Base   & 15.03 & 14.77 & 52.14 ($\uparrow$4.89\%) & 22.88 ($\uparrow$9.70\%) & 81.98 & 0.9689 \\ 
         &Corrector (Step1) & 15.03 & 14.72 & 55.6 ($\uparrow$7.67\%) & 25.54 ($\uparrow$12.20\%) & 84.85 & 0.9447 \\      
        & (Step 50+sbeam8)&	15.03 & 14.79& 	\underline{61.6} ($\uparrow$12.70\%)&	\textbf{31.39} ($\uparrow$25.33\%) &	86.67 &	0.9810 \\													
       \bottomrule
       
    \end{tabular}}
      \caption{Text Reconstruction within Language Family. For each language, the best TF1 is \underline{underlined}, and the best BLEU is \textbf{bolded}. The \colorbox{Dandelion}{color box} indicates the highest inversion boost compared to the baseline.}
 \label{tab:family-inversion}
\end{table}

\begin{table}[ht!]
    \centering
    \resizebox{\columnwidth}{!}{
  \begin{tabular}{c|l|c|c|c|c|c|c}
    \toprule
     & \textbf{Language}(Script)& \textbf{\#Tok.} & \textbf{\#Pred. Tok.} & \textbf{Token F1} & \textbf{BLEU} & \textbf{ROUGE} & \textbf{COS} \\ 
     \midrule
      \multirow{8}{*}{\rotatebox[origin=c]{90}{Gujarati-Kazakh}}
      &   \textbf{Gujarati} (Gujr) & ~ & ~ & ~ & ~ & ~ & ~ \\ 
        &Base & 11.07 & 10.82 & 45.34 ($\uparrow$4.98\%) & 19.75 ($\uparrow$5.39\%) & 88.21 & 0.9854 \\ 
        & Corrector(Step1) & 11.07 & 10.68 & 49.03 ($\uparrow$7.73\%) & 22.94 ($\uparrow$13.85\%) & 90.25 & 0.9804 \\ 
& (Step 50+sbeam8)	 &11.07	  &10.77 &	\underline{54.98} ($\uparrow$14.14\%)	& \textbf{29.79} ($\uparrow$33.27\%)	& 91.38 &	0.9809	\\									
        
       & \textbf{Kazakh} (Cyrl) & ~ & ~ & ~ & ~ & ~ & ~ \\ 
         &Base  & 13.35 & 13.21 & 49.34 ($\downarrow$7.12\%) & 26.08 ($\downarrow$15.02\%) & 85.22 & 0.9767 \\ 
         & Corrector(Step1) & 13.35 & 13.09 & 53.47 ($\downarrow$6.94\%) & 30.31 ($\downarrow$12.45\%) & 87.05 & 0.9800 \\ 
           & (Step 50+sbeam8) &	13.35&	13.14	& \underline{60.4} ($\downarrow$8.12\%) &	\textbf{38.20} ($\downarrow$14.16\%) &	89.89	& 0.9841 \\		
        \midrule

   \multirow{8}{*}{\rotatebox[origin=c]{90}{Hindi-Kazakh}}
     &    \textbf{Hindi} (Deva) & ~ & ~ & ~ & ~ & ~ & ~ \\ 
        &Base  & 15.21 & 14.97 & 52.16($\downarrow$2.31\%) & 19.58($\downarrow$2.51\%) & 74.22 & 0.9441 \\
        & Corrector(Step1) & 15.21 & 14.74 & 44.82 ($\downarrow$15.08\%) & 15.00 ($\downarrow$25.22\%) & 68.00 & 0.9584 \\ 
         & (Step 50+sbeam8) & 15.21 & 15.06 & \underline{56.96} ($\uparrow$2.10\%) & \textbf{23.99} ($\uparrow$7.63\%) & 76.15 & 0.9502 \\ 
         
       & \textbf{Kazakh} (Cyrl) & ~ &  ~ & ~ & ~ & ~ & ~ \\ 
        &Base  & 13.35 & 13.10 & 48.01 ($\downarrow$9.62\%) & 25.31 ($\downarrow$17.53\%) & 84.23 & 0.9563 \\ 
        & Corrector(Step1)  & 13.35 & 13.02 & 52.68 ($\downarrow$8.32\%) & 30.18 ($\downarrow$12.82\%) & 87.54 & 0.9729 \\
        & (Step 50+sbeam8)& 13.35 & 13.10 & \underline{59.38} ($\downarrow$9.67\%) & \textbf{36.57} ($\downarrow$17.81\%) & 89.07 & 0.9635 \\ 
        \midrule
        
\multirow{8}{*}{\rotatebox[origin=c]{90}{Kazakh-Punjabi}}
        
        & \textbf{Kazakh} (Cyrl) & ~ & ~ & ~ & ~ & ~ & ~ \\ 
       &Base & 13.35 & 13.06 & 48.34 ($\downarrow$9.00\%) & 25.60 ($\downarrow$16.59\%) & 84.63 & 0.9609 \\ 
        & Corrector(Step1) &	13.35&	13.09&	51.49 ($\downarrow$10.39\%) &	28.79 ($\downarrow$16.83\%) &	84.97	& 0.9662\\												
 & (Step 50+sbeam8) &	13.35 &	13.04 &	\underline{56.8} ($\downarrow$13.60\%)  &	\textbf{34.49} ($\downarrow$22.49\%)  &	86.12	 &0.9824\\															

      &  \textbf{Punjabi} (Guru) & ~ & ~ & ~ & ~ & ~ & ~ \\ 
        &Base   & 11.07 & 11.04 & 63.44 ($\uparrow$6.35\%) & 34.30 ($\uparrow$10.29\%) & 92.94 & 0.9880 \\ 
         & Corrector(Step1) & 11.07 & 10.93 &  65.92 ($\uparrow$$\uparrow$) & 37.28 ($\uparrow$$\uparrow$)& 93.89 & 0.9901 \\ 
          & (Step 50+sbeam8) & 	11.07&	10.95 &	\underline{71.67} ($\uparrow$$\uparrow$) & \textbf{46.65} ($\uparrow$$\uparrow$) &	95.19 &	0.9942	\\														

\midrule
        
        \multirow{8}{*}{\rotatebox[origin=c]{90}{Kazakh-Urdu}}  
       & \textbf{Kazakh} (Cyrl) & ~ & ~ & ~ & ~ & ~ & ~ \\ 
       &Base & 13.35 & 13.11 & 46.87 ($\downarrow$11.77\%)& 24.84 ($\downarrow$19.06\%) & 84.18 & 0.9619 \\
        & Corrector(Step1) & 13.35 & 21.00 & 0.2 ($\downarrow$99.65\%) & 0.03 ($\downarrow$99.91\%) & 0.00 & 0.7604 \\ 
        & (Step 50+sbeam8) &	13.35	 &21.00  &	\underline{0.2}	($\downarrow$99.70\%) &	\textbf{0.03} ($\downarrow$99.91\%) &	0 &	0.7528	\\															
        & \textbf{Urdu} (Arab) & ~ & ~ & ~ & ~ & ~ & ~ \\ 
        &Base & 15.03 & 14.81 & \underline{52.35} ($\uparrow$5.31\%) & \textbf{22.58} ($\uparrow$8.25\%) & 83.44 & 0.9657 \\ 
        & Corrector(Step1) & 15.03 & 21.00 & 0.08 ($\downarrow$99.85\%) & 0.01 ($\downarrow$99.96\%) &0 & 0.7687 \\ 
        & (Step 50+sbeam8) &	15.03	&21.00 &	0.08 ($\downarrow$99.84\%) &	0.01 ($\downarrow$99.96\%)		&0 & 0.7821	\\			

        \midrule

\multirow{8}{*}{\rotatebox[origin=c]{90}{Gujarati-Turkish}} 
        & \textbf{Gujarati (Gujr)} & ~ & ~ & ~ & ~ & ~ & ~ \\ 
       &Base & 11.07 & 10.75 & 42.64 ($\downarrow$1.27\%) & 18.84 ($\uparrow$0.53\%) & 87.55 & 0.9838 \\ 
          & Corrector(Step1) & 11.07 & 10.74 & 47.51 ($\uparrow$4.39\%) & 21.70 ($\uparrow$7.69\%) & 89.16 & 0.9849 \\ 
        & (Step 50+sbeam8) & 11.07 & 10.86 & \underline{54.49} ($\uparrow$13.12\%) & \textbf{28.19} ($\uparrow$26.13\%) & 91.61 & 0.9892 \\ 
        
        & \textbf{Turkish (Latn)} & ~ & ~ & ~ & ~ & ~ & ~ \\ 
       &Base & 14.66 & 14.42 & 41.08 ($\downarrow$6.74\%) & 16.10 ($\downarrow$11.29\%) & 46.29 & 0.9437 \\ 
          & Corrector(Step1)  &	14.66	&14.37	&45.07	($\downarrow$2.72\%) &	19.37 ($\downarrow$6.04\%) &	51.09	& 0.9657	\\															
 & (Step 50+sbeam8)	 &	14.66  &		14.44 &		\underline{50.17} ($\downarrow$0.42\%)	 &	\textbf{24.29} ($\downarrow$0.91\%)	 &	57.46  & 0.9593	\\														

\midrule

         \multirow{8}{*}{\rotatebox[origin=c]{90}{Hindi-Turkish}}

         & \textbf{Hindi} (Deva) & ~ & ~ & ~ & ~ & ~ & ~ \\ 
       &Base & 15.21 & 15.00 & 53.57 ($\uparrow$5.08\%) & 20.65 ($\uparrow$8.12\%) & 76.03 & 0.9518 \\ 
       & Corrector(Step1) & 15.21 & 14.88 & 55.97 ($\uparrow$6.04\%) & 23.05 ($\uparrow$14.91\%) & 77.09 & 0.9743 \\ 
       & (Step 50+sbeam8) & 	15.21	&15.00	& \underline{61.34} ($\uparrow$9.95\%) &	\textbf{28.59} ($\uparrow$28.26\%) & 80.74	&  0.9917	\\											
       
         & \textbf{Turkish} (Latn) & ~ & ~ & ~ & ~ & ~ & ~ \\ 
        &Base  & 14.66 & 14.49 & 42.85 ($\downarrow$2.72\%) & 17.11 ($\downarrow$5.73\%) & 48.13 & 0.9413 \\ 
        & Corrector(Step1) & 14.66 & 14.26 & 44.69 ($\downarrow$3.54\%) & 19.50 ($\downarrow$5.43\%) & 51.37 & 0.9738 \\ 
     & (Step 50+sbeam8) &	14.66	 &14.35	 & \underline{51.08} ($\uparrow$10.25\%) &	\textbf{25.58} ($\uparrow$24.10\%) &	58.27 &	0.9498\\																										

\midrule

  \multirow{8}{*}{\rotatebox[origin=c]{90}{Punjabi-Turkish}} 
        & \textbf{Punjabi} (Guru) & ~ & ~ & ~ & ~ & ~ & ~ \\ 
        &Base  & 11.07 & 10.91 & \underline{62.93} ($\uparrow$5.50\%) & \textbf{34.19} ($\uparrow$9.94\%) & 92.68 & 0.9878 \\ 
        & Corrector(Step1)  & 11.07 & 10.56 & 38.58 ($\uparrow$$\uparrow$) & 13.69 ($\uparrow$$\uparrow$) & 88.89 & 0.9711 \\ 
         & (Step 50+sbeam8) &	11.07	 &10.87	 & 54.33 ($\uparrow$$\uparrow$) &		25.37 ($\uparrow$$\uparrow$)		 &	92.28 &	0.9904	\\							
        
        & \textbf{Turkish} (Latn) & ~ & ~ & ~ & ~ & ~ & ~ \\ 
        &Base  & 14.66 & 14.28 & 41.97 ($\downarrow$4.72\%)& 16.75 ( $\downarrow$7.71\%) & 47.29 & 0.9298 \\ 
      
          & Corrector(Step1) & 	14.66	&14.31	& 44.25 ($\downarrow$4.49\%) &	18.89 ($\downarrow$8.36\%) &	50.71	&0.9445	\\																		
 & (Step 50+sbeam8)  &	14.66 &	14.45	& \underline{49.2} ($\downarrow$2.34\%) &	\textbf{23.52} ($\downarrow$4.04\%) &	56.18 &	0.9463	\\																		

\midrule

     \multirow{8}{*}{\rotatebox[origin=c]{90}{Turkish-Urdu}} 

       & \textbf{Turkish} (Latn) & ~ & ~ & ~ & ~ & ~ & ~ \\ 
      &Base & 14.66 & 14.49 & 41.65 ($\downarrow$5.45\%) & 16.70 ($\downarrow$7.99\%) & 47.32 & 0.9373 \\ 
& Corrector(Step1) & 	14.66	 &14.34	 &44.93 ($\downarrow$3.02\%)  &	19.32 ($\downarrow$6.30\%)  &	51.26	 &0.9631	\\			
& (Step 50+sbeam8) &	14.66	&14.39	& \underline{50.05} ($\downarrow$0.66\%) &	\textbf{23.95} ($\downarrow$2.28\%) &	56.61 &	0.9083	\\																		

       &  \textbf{Urdu} (Arab) & ~ & ~ & ~ & ~ & ~ & ~ \\ 
       & Base & 15.03 & 14.74 & 53.79 ($\uparrow$8.21\%) & 23.51 ($\uparrow$12.70\%) & 83.57 & 0.9574 \\ 
      &Corrector(Step1) & 15.03 & 14.70 & 56.03 ($\uparrow$8.50\%) & 26.27 ($\uparrow$15.42\%) & 85.51 & 0.9632 \\ 
     & (Step 50+sbeam8)  & 15.03 & 14.86 & \underline{61.6} ($\uparrow$12.70\%) & \textbf{31.77} ($\uparrow$26.88\%)& 87.82 & 0.9841 \\ 
     \bottomrule
    \end{tabular}}
    \caption{Text Reconstruction in Control Group. For each language, the best TF1 is \underline{underlined}, the best BLEU is \textbf{bolded}. }
 \label{tab:random-inversion}
\end{table}

\begin{figure*}[!ht]
\centering
\includegraphics[width=0.9\textwidth]{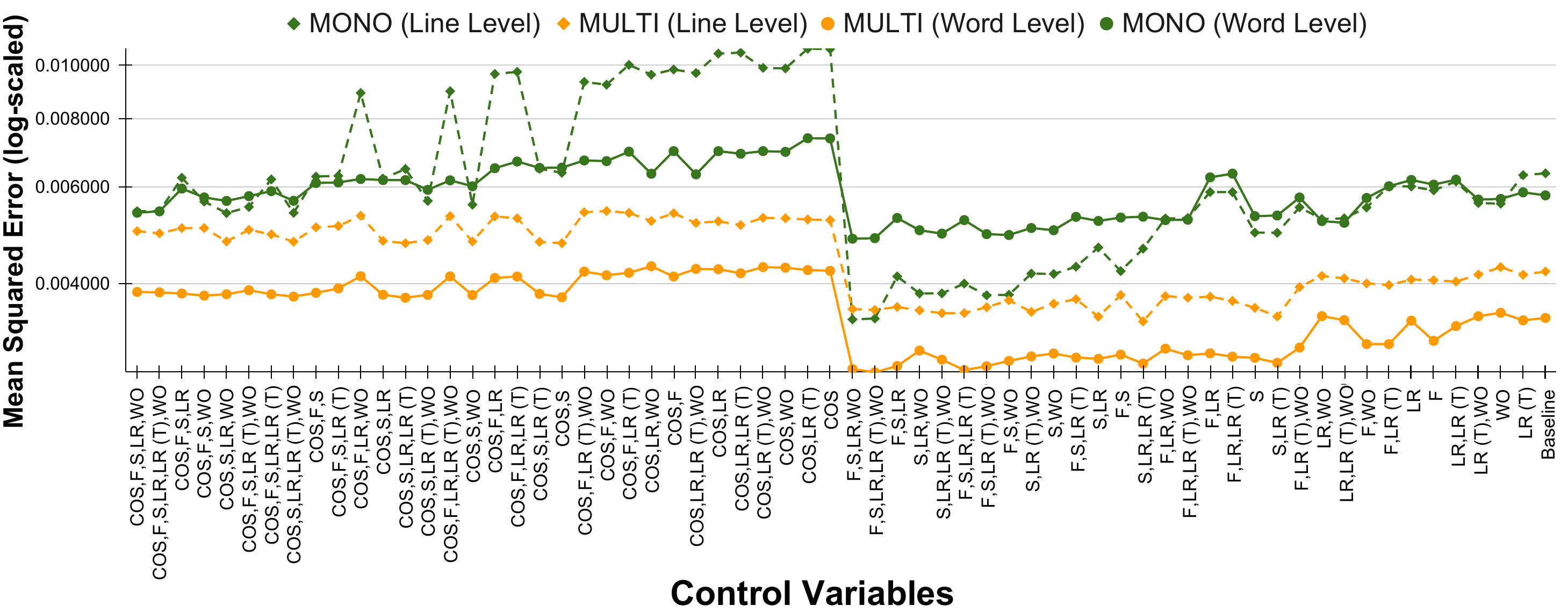}
\caption{Language Confusion in inverting \textcolor{teal}{Monolingual (\textsc{mono})} and \textcolor{orange}{Multilingual (\textsc{multi})} textual embeddings with Control Variables from Linguistic Characteristics.}
\label{fig:regression_results}
\end{figure*}

\paragraph{Attacking In-Family Text Embeddings}
In addition to script being a significant factor in enhancing inversion attack performance, we also investigate the impact of a shared language family in this context. We train and evaluate various language pairs selected from the Semitic, Turkic, and Indo-Aryan language families, as shown in Table~\ref{tab:family-inversion}. 
Overall, inversion attack performance in word-matching metrics improves for all languages across evaluation steps, with the exception of Gujarati at the corrector step1 and Hindi with the base model. Notably, the corrector model trained solely on Punjabi fails to produce meaningful output when inverting Punjabi text embeddings in the baseline scenario (ref. Table~\ref{tab:baseline-results}). However, when trained alongside other Indo-Aryan languages, the performance improves dramatically. For instance, when Punjabi is trained with Gujarati, the BLEU score for the corrector model at step 50 with a beam search width of 8 jumps from 0 to 52.89.
Additionally, Turkish text embedding inversion performs better in in-family training compared to in-script training (ref. Table~\ref{tab:in-script-results}). 
Overall, training the embedding inversion model with languages from the same language family significantly enhances performance.
This indicates the vulnerability of low-resource languages which are related to high-resource ones.

\paragraph{Control Group in Text Embedding Inversion}
To rule out the possibility that performance gains are merely due to an increased amount of training data compared to baseline models, we conduct experiments with a control group where languages from the Indo-Aryan and Turkic families are mixed. 
As shown in Table~\ref{tab:random-inversion}, when paired with Kazakh, the inversion performance of Indo-Aryan languages improves compared to the baseline, with the exception of Urdu when using corrector models. However, the improvement is not as significant as that observed with in-family training (ref. Table~\ref{tab:family-inversion}). Conversely, the performance of inverting Kazakh text embeddings diminishes with these combinations.
A similar trend is observed in inversion models trained on combinations of Turkish and Indo-Aryan languages, except when using the combination of Turkish and Hindi text embeddings. In this case, the reconstruction of Turkish texts also improves performance in word-matching metrics, possibly due to overlapping vocabulary between Hindi and Turkish~\citep{malaiya_turkish}.
Overall, this rules out the size of the training data as the decisive factor in enhancing inversion attack performance; instead, shared script and language family are necessary contributors.

\section{Language Confusion Analysis}
When \textsc{gtr} is used for the black-box encoder with German training data as depicted in Fig.~\ref{fig:language_confusion_mono_multi_levels} (left), the output languages at both word and line levels are a mix of German, English, and other languages, with proportions varying across different steps. However, with \textsc{me5} as the black-box encoder, the output is more likely to remain in the source language (i.e., German) at both levels.
We also observe the target and hypothesis embeddings of inverted Chinese texts across different stages using inversion models trained on monolingual and multilingual embeddings in German. 
For multilingual embeddings, the target and hypothesis embeddings form distinct clusters, consistent with the findings that multilingual embeddings in different languages tend to occupy their respective subspaces.
In contrast, for monolingual embeddings, the hypothesis embeddings are clustered together during inference steps, while at the base model stage, they are more sparsely distributed (see details in SUPPL.).

\paragraph{Patterns of Language Confusion}
This phenomenon of language confusion is pervasive and seems to be random at first. 
However, inversion attack results show there are strong correlations between the improved performance with training models using languages with shared scripts and language families. 
The language of training data, evaluation data, and different evaluation stages may serve as baseline variables for predicting language patterns, acting as control variables for the baseline model.
In addition, shared language family and scripts and also the cosine similarity of the target and hypothesis embeddings play a role in language confusion. 
We use the combinations of COS, F, S, LR, LR(T), WO plus the variables from baseline, to predict the language of the inversion output using Random Forest Regression, with Mean Squared Error (MSE) as evaluation metrics.
As shown in Fig.~\ref{fig:regression_results}, all combinations of control variables with COS predict languages with larger error for both monolingual and multilingual LLMs and also for both word and line levels, while consistently, the features such as F, S, LR, LR(T) and WO renders better predictions across LLMs and levels.

\section{Conclusion}

%

Our work highlights the fact that vulnerabilities in multilingual embedding attacks disproportionately affect low-resource languages.
This risk is further exacerbated by the fact that language confusion, which can be exploited by attackers, is shown to be both predictable and solvable. 
We prove this by showing directly that attackers drastically increase the efficacy of their embedding inversion attacks, when specifically accounting for language confusion (i.e., by training on in-family and in-script languages), making it an appealing exploit for malicious actors to potentially leverage. 
In this context, we argue that a deeper understanding of language confusion, including its connection to linguistic characteristics, could offer critical insights into the mechanics of multilingual attacks.
Ultimately, the security of LLMs and guarantees for safety for all end-users relies on the inclusion of diverse languages in LLM security research.
Multilinguality poses unique challenges, and solutions offered for high-resource languages are not necessarily useful for those with fewer resources.
Without prioritizing diversity in this emerging field, LLMs will remain vulnerable, with marginalized communities worldwide facing disproportionate risk as LLMs become more widespread.

\section*{Limitations}
\paragraph{Computational Constraints}
Despite reducing the training data size for each inversion model by 60\% to 80\%, allowing us to investigate significantly more languages compared to previous studies~\citep{morris-etal-2023-text,chen2024text}, the computational overhead remains substantial. 
A key limitation of this work is the resource-intensive nature of the experiments, which required approximately 40,000 GPU hours. 
While expanding this research to include additional languages would further escalate computational costs, we strongly advocate prioritizing NLP security across diverse languages to ensure equitable advancements beyond English.


\paragraph{Defense Mechanisms}
This work investigates multilingual inversion attacks in a wide range of languages and exposes patterns of language confusion in these settings. 
However, the inversion methodology is adopted from~\citet{chen2024text}, and the defense mechanism, the so-called masking defense, proposed in their work also applies to our setting. 

\paragraph{LLM's Cross-lingual Vulnerabilities}
Defending multilingual LLMs has proven particularly challenging against attacks such as jailbreak~\citep{he2024tuba}, backdoor attacks~\citep{wang2024backdoor}, and fine-tuning~\citep{poppi2024towards}. 
We echo these findings by examining multilingual LLMs in embedding inversion attacks across a linguistically diverse set of languages. 
A major finding of this work is that language confusion can be directly weaponized as an exploit against multilingual LLMs, and as such, we provide further analysis to better understand the underlying patterns of language confusion. 
Specifically, we present the relations between vulnerabilities and linguistic characteristics to advocate a more linguistically nuanced approach to understanding the vulnerability for linguistically guided mitigation and defense in future research. 
For example, as indicated in Table~\ref{tab:random-inversion} and Fig.~\ref{fig:regression_results}, LLMs trained on language pairs from different scripts and families (specifically for Kazakh and Urdu, or Hindi and Kazakh)  are more robust against inversion attacks and language confusion. However, the trade-off between LLM vulnerabilities and LLM performance still needs to be investigated thoroughly in multilingual settings to devise effective defense mechanisms, which we leave for future work.


\section*{Ethics Statement}
This work investigates attacks on multilingual embedding models and the phenomenon of language confusion, with the goal of highlighting vulnerabilities across languages representing diverse scripts and typologies.
We encourage the community to incorporate a broader range of languages in NLP security research.
Some of the languages we examine, such as Yiddish and Meadow Mari, are particularly vulnerable. 
To minimize potential harm, \emph{we do not train any attack models} on these languages, instead focusing solely on evaluation to determine their current threat level. 
The language models used in this study are open-source, ensuring that this research does not pose an immediate threat to EaaS providers, who are likely using private models. 
Additionally, we take care not to experiment with any sensitive data, ensuring that no real-world harm results from this work.

\section*{Acknowledgements}
YC, HCL, and JB are funded by the Carlsberg Foundation, under the Semper Ardens: Accelerate programme (project nr.~CF21-0454). 
We further acknowledge the support of the AAU AI Cloud and express our gratitude to DeiC for providing computing resources on the LUMI cluster (project no. DeiC-AAU-N5-2024085-H2-2024-28). 
We extend our thanks to Sighvatur Sveinn Davidsson and Frederik Petri Svenningsen for their assistance with AI Cloud, and to Esther Ploeger and Mike Zhang for their discussions related to this work.

\bibliography{aaai24}

\newpage
\appendix

\begin{table*}[!ht]
    \centering
     \resizebox{\textwidth}{!}{
    \begin{tabular}{cccccccc}
    \toprule
        Language & ISO 639 & Lang. Family & Lang. Script & Script ISO & Directionality & \#Samples(Train) & WO \\ 
        \midrule
        Arabic & arb & Semitic & Arabic & Arab & RTL & 1M & VSO \\ 
        Urdu & urd & Indo-Aryan & Arabic & Arab & RTL & 600K & SOV \\ 
        Kazakh & kaz & Turkic & Cyrillic & Cyrl & LTR & 1M & SOV \\ 
        Mongolian & mon & Mongolic & Cyrillic & Cyrl & LTR & 1M  & SOV \\ 
        Hindi & hin & Indo-Aryan & Devanagari & Deva & LTR  &  600K & SOV \\ 
        Gujarati & guj & Indo-Aryan & Gujarati & Gujr & LTR &  600K & SOV \\ 
        Punjabi & pan & Indo-Aryan & Gurmukhi & Guru & LTR &  600K & SOV \\ 
        Chinese & cmn & Sino-Tibetan & Haqniqdoq & Hani & LTR & 1M & SVO\\ 
        Hebrew & heb & Semitic & Hebrew & Hebr & RTL & 1M & SVO\\ 
        Japanese & jpn & Japonic & Japanese & Jpan & LTR & 1M & SOV\\ 
        German & deu & Germanic & Latin & Latn & LTR & 1M  & Non-Dominant\\ 
        Turkish & tur & Turkic & Latin & Latn & LTR & 1M & SOV \\ 
        Amharic & amh & Semitic & Ethiopian & Ethi & LTR &  - & SOV \\ 
        Sinhala & sin & Indo-Aryan & Sinhala & Sinh & LTR  & -& SOV \\ 
        Korean & kor & Koreanic & Hangul & Hang & LTR & - & SOV \\ 
        Finnish & fin & Uralic & Latin & Latn & LTR & - & SVO \\ 
        Hungarian & hun & Uralic & Latin & Latn & LTR & -  & Non-Dominant \\ 
        Yiddish & ydd & Germanic & Hebrew & Hebr & RTL & - & SVO \\ 
        Maltese & mlt & Semitic & Latin & Latn & LTR & - & Non-Dominant \\ 
        Meadow Mari & mhr & Uralic & Cyrillic & Cyrl & LTR & - & SOV \\

        \bottomrule
    \end{tabular}}
    \caption{Languages and their Language Characteristics, i.e., Language Family, Language Script, Directionality of the Script, Number of Training Samples for Inversion Models, Word Order of Subject, Object and Verb.  }
    \label{tab:languages}
\end{table*}

\section{Features for  Language Confusion}
To model and predict patterns of language confusion in the context of multilingual inversion attacks, we leverage the linguistic characteristics of both the evaluation languages and the train data used for the corresponding inversion models. 
Consider an inversion model trained on a set of \( n \in \mathbf{N} \) languages, denoted as \( L_s = \{l_1, \dots, l_n\} \). If the evaluation dataset is in a language \( l_t \), and the languages listed in Table~\ref{tab:languages} form a non-empty set \( L \), then it follows that \( L_s \subseteq L \) and \( l_t \in L \).
For a linguistic feature \( f \) (e.g., language family, script, script directionality, or word order of subject, object, and verb), we assign a value of 1 if \( f_{l_t} \) matches \( f_{l_i} \) for any \( l_i \in L_s \); otherwise, we assign it a value of 0. 
These binary variables are then used as inputs for training the regression model.
We use `LabelEncoder'\footnote{\url{https://scikit-learn.org/stable/modules/generated/sklearn.preprocessing.LabelEncoder.html}} to encode the evaluation language \( l_t \), and `MultiLabelBinarizer'\footnote{\url{https://scikit-learn.org/stable/modules/generated/sklearn.preprocessing.MultiLabelBinarizer.html}} to encode the set of training languages \( L_s \). 
The evaluation stages are handled using `OneHotEncoder'\footnote{\url{https://scikit-learn.org/stable/modules/generated/sklearn.preprocessing.OneHotEncoder.html}}. 
Cosine similarities calculated for the evaluation are directly incorporated as one of the features. 
The probabilities of decoded languages, whether at the word level or line level, are used directly as the target variables.

\section{Analyzed LLMs}
Table~\ref{tab:llms} shows the details of the pre-trained LLMs used in this work, including the pre-training dataset, the number of languages supported, the model architecture, and whether the model consists of an encoder, a decoder or both.

\begin{table*}[ht!]
     \centering
     \resizebox{\textwidth}{!}{
    \begin{tabular}{ccccc}
    \toprule
    \textbf{Model} & \textbf{\#Lang.} &  \textbf{Dataset} & \textbf{Encoder/Decoder} & \textbf{Architecture }\\
    \midrule
    \textsc{T5} & 4 & C4 & encoder-decoder & T5 \\
    \textsc{mT5} & 102 & mC4 & encoder-decoder & T5 \\
    \textsc{me5} & 102 & Wikipedia & encoder  & \textsc{xlm-r}\\
    \textsc{gtr}& 1 (English) & MS Marco~\citep{nguyen2016ms}, Community QA, Natural Questions~\citep{kwiatkowski-etal-2019-natural} & encoder &  \textsc{t5} \\
   \textsc{alephbert} & 1 (Hebrew) & Oscar~\citep{ortiz-suarez-etal-2020-monolingual}, Wikipedia, Twitter & encoder & \textsc{bert} \\
    \textsc{text2vec} & 1 (Chinese) & NLI data in Chinese~\citep{cmn_text2vec_data} & encoder & \textsc{CoSENT} \\
    
    \bottomrule

    \end{tabular}}
    \caption{Details of pre-trained LLMs used in this work. }
    \label{tab:llms}
\end{table*}

\section{Visualizing Embeddings}

\begin{figure*}[th!]
\centering
\includegraphics[width=\textwidth]{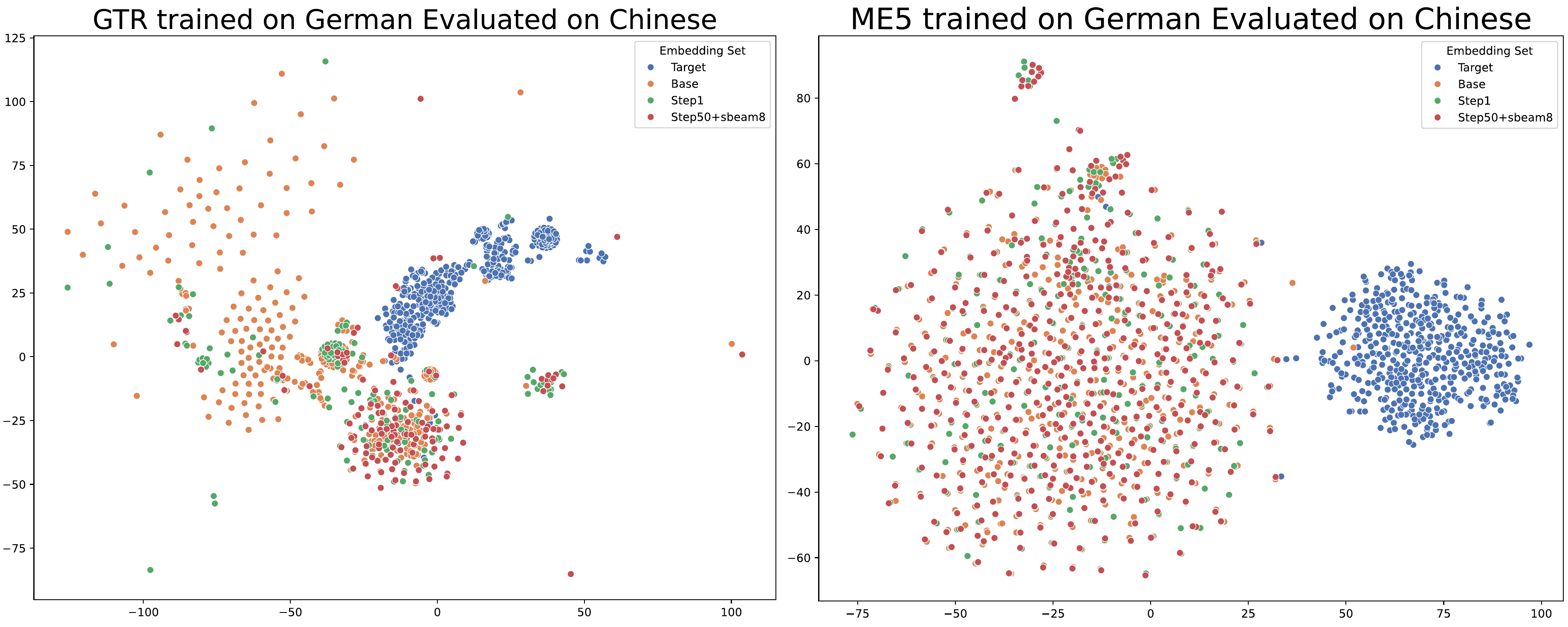}

\caption{Embeddings of reconstructing Chinese (Hani) texts using inversion models trained on Monolingual (left) and Multilingual (right) encoders with German (Latn).}
\label{fig:embeddings_multi_mono}
\end{figure*}

As illustrated in Fig.~\ref{fig:embeddings_multi_mono}, we observe the target and hypothesis embeddings of inverted Chinese texts across different stages using inversion models trained on monolingual and multilingual embeddings in German. 
For multilingual embeddings, the target and hypothesis embeddings form distinct clusters, consistent with the findings that multilingual embeddings in different languages tend to occupy their respective subspaces.
In contrast, for monolingual embeddings, the hypothesis embeddings are clustered together during corrector steps, while at the base model stage, they are more sparsely distributed.

\end{document}